\pdfoutput=1
\PassOptionsToPackage{table}{xcolor}
\documentclass[11pt]{article}

\usepackage[final]{acl}

\usepackage[table]{xcolor} 
\usepackage{times}
\usepackage{latexsym}
\usepackage{algorithm}
\usepackage{algpseudocode}
\usepackage{amsmath}
\usepackage[utf8]{inputenc}
\usepackage{placeins}
\usepackage{booktabs} 
\definecolor{myGreen}{HTML}{C6EFCE} 
\definecolor{myRed}{HTML}{FFC7CE}   
\definecolor{myYellow}{HTML}{F7F2A6}

\usepackage{lipsum}
\usepackage{graphicx}
\usepackage{amsmath}
\usepackage{etoolbox,xspace}
\usepackage{pdfpages}
\usepackage{breqn}
\usepackage[most]{tcolorbox}
\usepackage{fontawesome}
\usepackage{enumitem}
\usepackage{multirow} 
 \usepackage{amssymb}

\newcommand*{\llama}{\textit{LLaMA}}

\newcommand*{\gemma}{\textit{Gemma}}
\newcommand*{\llamathree}{\textit{LLaMA-3-8B-Instruct}}
\newcommand*{\gemmatsit}{\textit{Gemma-2-27b-it}}
\newcommand*{\llamasit}{\textit{LLaMA-3-70B-Instruct}}

\newcommand*{\gemmatwentyit}{\textit{Gemma-2-27b-it}}
\newcommand*{\gemmanineit}{\textit{Gemma-2-9b-it}}
\newcommand*{\llamathreeit}{\textit{LLaMA-3-8B-Instruct}}
\newcommand*{\cohere}{\textit{Aya-expanse-32b}}
\newcommand*{\deepseek}{\textit{Deepseek-r1-distill-llama-70b}}
\newcommand*{\nemo}{\textit{Mistrial-nemo-instruct-2407}}

\newcommand*{\geminionepro}{\textit{Gemini-1.5-pro}}
\usepackage[T1]{fontenc}
\newcommand*{\geminitwopro}{\textit{Gemini-2.5-pro}}
\newcommand*{\gptfmini}{\textit{GPT-4o-mini}}

\newcommand\blfootnote[1]{%
  \begingroup
  \renewcommand\thefootnote{}\footnote{#1}%
  \addtocounter{footnote}{-1}%
  \endgroup
}

\usepackage[T1]{fontenc}

\usepackage[utf8]{inputenc}

\usepackage{microtype}
\usepackage{booktabs}

\usepackage{tikz}
\newcommand*\circled[1]{\tikz[baseline=(char.base)]{
            \node[shape=circle,draw,inner sep=1pt] (char) {#1};}}

\usepackage{inconsolata}

\usepackage{graphicx}

\title{\textit{Bridging the Culture Gap}: A Framework for LLM-Driven Socio-Cultural Localization of Math Word Problems in Low-Resource Languages}

\author{\normalsize Israel Abebe Azime$^{1,\ast }$,  Tadesse Destaw Belay$^{2,\ast}$, Dietrich Klakow$^{1}$, Philipp Slusallek$^{1}$ \\ 
\textbf{\normalsize Anshuman Chhabra$^{3}$ } \\
\footnotesize
 $^1$ Saarland University, Saarland Informatics Campus, Germany  $^2$ Instituto Politécnico Nacional, Mexico
\\ 
 \footnotesize
    $^3$ University of South Florida, Tampa, FL USA
\\}

\begin{document}
\maketitle
\blfootnote{$^\ast$ Equal Contribution.}
\begin{abstract}
Large language models (LLMs) have demonstrated significant capabilities in solving mathematical problems expressed in natural language. However, multilingual and culturally-grounded mathematical reasoning in low-resource languages lags behind English due to the scarcity of socio-cultural task datasets that reflect accurate native entities such as person names, organization names, and currencies. Existing multilingual benchmarks are predominantly produced via translation and typically retain English-centric entities, owing to the high cost associated with human annotater-based localization. Moreover, automated localization tools are limited, and hence, truly localized datasets remain scarce. In this work, we study the cultural robustness of large language models by examining the impact of culturally specific entities and the biases introduced by English-centric benchmarks. To bridge this gap, we introduce a framework for LLM-driven cultural localization of math word problems that automatically constructs datasets with native names, organizations, and currencies from existing sources. We find that translated benchmarks can obscure true multilingual math ability under appropriate socio-cultural contexts. Through extensive experiments, we also show that our framework can help mitigate English-centric entity bias and improves robustness when native entities are introduced across various languages.
\end{abstract}

\section{Introduction}
\begin{figure}[!h]
\centering
\includegraphics[width=0.49\textwidth]{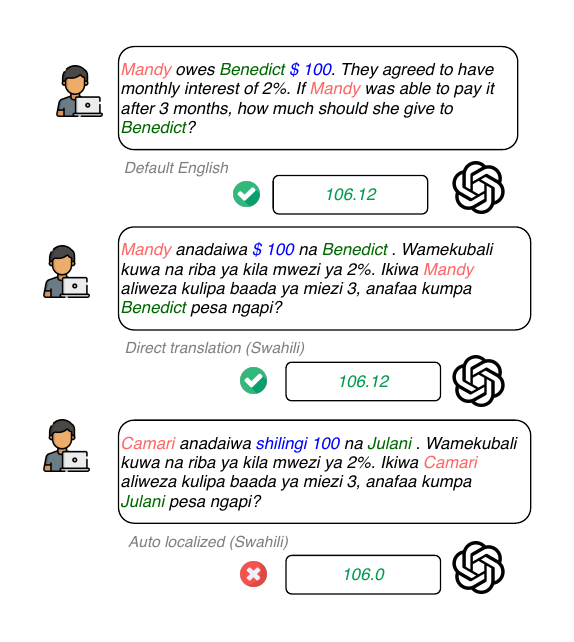}\vspace{-5mm}
\caption{An example showcasing an English math word problem, its direct translation, and an automatically localized version with culturally adapted entities. While the problem structure remains identical, large language models (LLMs) often fail to answer correctly when entity names or currencies are altered. This highlights a key limitation in current LLM robustness. In this paper, our goal is to audit models and rectify their robustness issues so that remain consistent and accurate across such culturally grounded variations.}\label{fig:intro}
\end{figure}


Mathematical reasoning has been adopted as a core milestone in large language model understanding research~\cite{yan2024survey,ahn2024large,ahn-etal-2024-large}. Math word problems (MWPs), a key component of mathematical reasoning tasks, have been widely explored as a challenging benchmark for LLMs~\cite{srivatsa-kochmar-2024-makes}. MWPs are characterized by their \textit{integration of mathematical knowledge with scenarios drawn from everyday activities, which can vary across cultures}. The ability of low-resource and multilingual LLMs to solve MWPs correctly depends on multiple factors: (1) the language used to prompt the models~\cite{adelani2024irokobench} and (2) the linguistic complexity of the questions~\cite{srivatsa-kochmar-2024-makes}. Despite their impressive performance on MWPs, a crucial question remains: \textit{do LLMs truly retain performance under culturally diverse MWPs, even when the underlying mathematical structure remains unchanged?}
Math word problems are not culture-free \cite{tomar2025mathematicsisntculturefree}, incorporating cultural elements such as personal names, organizational names, currencies, measurement units, and others. The effects of these cultural element changes in math word problems across languages are underexplored.

Low-resource language (LRL) math word problem evaluation heavily relies on human or machine-translated benchmarks, as demonstrated by the translation of GSM8K~\cite{cobbe2021training} (a dataset of 8.79K high-quality, linguistically diverse grade school math word problems) into GSM8K\_zh for Chinese~\cite{yu2023metamath}, MGSM for 10 typologically diverse languages~\cite{shi2022multilingualchainofthought}, and AfriMGSM for 18 African languages~\cite{adelani2024irokobench}. Existing human- or machine-translated benchmarks often fail to account for local cultural contexts, which include day-to-day activities, common names, and local currencies. As a result, evaluating LLMs on such non-localized data primarily measures their mathematical understanding ability in English-centric scenarios, names, and currencies.

In this work, we investigate the task of creating more culturally aligned math word problems from their translated variants. Owing to the high costs associated with human-based localization, it is imperative to develop automated frameworks capable of producing large-scale datasets. Therefore, this motivates our first research question in this work: (\textbf{RQ1}) \textit{Can LLM-driven pipelines be used to localize translated benchmarks for low-resource languages?}

Moreover, given that a significant number of benchmark datasets are created using translation ~\cite{alabi2025charting} and do not study appropriate socio-culturally grounded datasets (see Figure \ref{fig:intro}) we propose an automated framework for socio-cultural localization of MWPs in low-resource languages. Using our framework, we can study our second research question, as follows: (\textbf{RQ2}) \textit{Does introducing socio-cultural local entities in translated existing benchmarks reveal performance disparities?} We undertake several experiments to study RQ2 for both low-resource languages (LRLs) and English to ascertain the performance gaps. Finally, we utilize our framework to generate localized MWPs for 18 African Languages covered by~\cite{adelani2024irokobench} and investigate our final research question: (\textbf{RQ3}) \textit{Can automated localization improve model robustness by augmenting benchmarks with culturally adapted variants?}

To address these questions, our work makes the following contributions:
\begin{itemize}

    \item We develop an LLM-driven localization pipeline to generate culturally adapted versions of translated datasets by replacing key entities with apt socio-cultural variants and processed localized MWP datasets.\footnote{ \url{https://github.com/IsraelAbebe/Auto-Localizer}}.
    \item Further, through extensive experiments enabled via our framework, we investigate performance disparities between localized and directly translated benchmarks across several LLMs like \gemma~and \llama~models.
    Our findings reveal how the presence of cultural entities influences LLMs' ability to solve mathematical word problems. For instance, we observe as much as 9\% (numeric match) drop in performance for \gptfmini \, (with similar trends seen across other LLMs). 
    \item Then, using the localized data generated by our automated localization framework, we fine-tune LLMs on these socio-cultural MWP variants and find that model robustness and generalization improves significantly across multiple languages, thereby paving the way for improving the performance of LLMs in a straightforward manner.
\end{itemize}

\begin{table*}[!ht]
\footnotesize
\centering
\begin{tabular}{l|p{10cm}}
\toprule
\textbf{Stages} & \textbf{Example } \\
\midrule
\texttt{\circled{1} English} $x_{\text{en}}$ & 

 \texttt{
 \textcolor{red}{Mandy} owes \textcolor{green}{Benedict} \textcolor{blue}{\$ 100}. They agreed to have monthly interest of 2\%. If \textcolor{red}{Mandy} was able to pay it after 3 months, how much should she give to \textcolor{green}{Benedict}?
 }
  \\
\midrule
\texttt{\circled{2} Translated   } \\
$x_{\text{trans}} \leftarrow x_{\text{en}}$ & \vspace{-0.5cm} \texttt{\textcolor{red}{Mandy} anadaiwa \textcolor{blue}{\$ 100} na \textcolor{green}{Benedict} . Wamekubali kuwa na riba ya kila mwezi ya 2\%. Ikiwa \textcolor{red}{Mandy} aliweza kulipa baada ya miezi 3, anafaa kumpa \textcolor{green}{Benedict}  pesa ngapi?
} \\
\midrule
\texttt{\circled{3} Important entities} \\$ent= \texttt{\colorbox{myYellow}{LLM}} (x_{\text{en}})$ & \vspace{-0.5cm}
\texttt{\{``personal\_names'':[``Mandy'',``Benedict''],``currencies'':[``\$'']'', ``organization\_names'':[]\}} \\
\midrule
\texttt{\circled{4} replacement dictionary   } \\ 
$replacement\_dict= fn (en, db)$ & \vspace{-0.5cm} \texttt{\{``Mandy'': ``Camari'', ``Benedict'': ``Julani'', ``\$'': ``shilingi'', ``dollar'': ``shilingi''\} 
} \\
\midrule
\texttt{\circled{5} Entity replaced } \\
$x_{\text{ent}}= fn (x_{\text{trans}}, replacement\_dict)$  & \vspace{-0.5cm}\texttt{\textcolor{red}{Camari} owes \textcolor{green}{Julani} \textcolor{blue}{shilingi 100}. They agreed to have monthly interest of 2\%. If \textcolor{red}{Camari} was able to pay it after 3 months, how much should she give to \textcolor{green}{Julani}?}\\
\midrule
\texttt{\circled{6} Auto Localized}\\
 \\ $\hat{x}_{\text{loc}}= \texttt{\colorbox{myYellow}{LLM}} ((x_{\text{en}}, x_{\text{trans}}), x_{\text{ent}})$  &  \vspace{-1cm}\texttt{\textcolor{red}{Camari} anadaiwa \textcolor{blue}{shilingi 100} na \textcolor{green}{Julani} . Wamekubali kuwa na riba ya kila mwezi ya 2\%. Ikiwa \textcolor{red}{Camari} aliweza kulipa baada ya miezi 3, anafaa kumpa \textcolor{green}{Julani}  pesa ngapi?}\\
\midrule
\texttt{\circled{7} Quality Check }\\
 &  \vspace{-0.5cm}\texttt{length($\hat{x}_{\text{loc}}$) == length($x_{\text{trans}}$), key entities not in $\hat{x}_{\text{loc}}$, replacement entities in $\hat{x}_{\text{loc}}$ , and similarity($\hat{x}_{\text{loc}}$, $x_{\text{trans}}$) > 0.8, Full human verification for the test set and sampled human verification for the training set. if failed return ($x_{\text{trans}}$) }\\
\bottomrule 
\end{tabular}
\caption{\textbf{The different stages of our MWP automated localization framework for low-resource languages}.
We show a step-by-step transformation from English to a culturally and linguistically localized Swahili version: direct translation, name and currency replacements, a semi-localized substitution, and the final fluent localization. Colored highlights indicate aligned entities across languages; occurrences of \colorbox{myYellow}{\texttt{LLM}} denote the use of an LLM.}
\label{stages}\vspace{-2mm}
\end{table*}

\section{Related Work}
\noindent\textbf{Math World Problems.}
Math word problems are mathematical questions framed in everyday language, often grounded in real-life scenarios and activities rather than expressed purely through abstract symbols or equations~\cite{cobbe2021training}. These problems serve as an engaging way to learn mathematical reasoning, as they require not only computational skills but also the ability to interpret and model real-world situations.

\noindent\textbf{LLM Robustness.} Although LLMs are highly capable of solving complex tasks, they often struggle with simple variations in input~\citep{askari2025assessing, chhabra2024revisiting, achara2025watching}. For example,~\citet{abedin2025arithmattack} demonstrate that even minor perturbations such as spelling mistakes can significantly impact model performance, a challenge that is particularly pronounced in low-resource settings and evaluations.

\noindent\textbf{Translation and Localization.} Translating existing benchmarks is one of the simplest approaches to benchmark creation, particularly for low-resource languages that lack dedicated evaluation datasets. In such cases, researchers often resort to translating established benchmarks from high-resource languages to enable evaluation and comparison~\cite{adelani2024irokobench,koto2024arabicmmlu,li2023cmmlu,son2024kmmlu}. In African context close to 30\% of resource papers translate existing benchmarks to create language specific benchmarks~\cite{alabi2025charting}.  However, to better reflect native cultural identities and move away from the Western-centric concepts that often persist in translated benchmarks, some researchers have begun developing fully localized benchmarks~\cite{yu2025injongo}. This work complements manual localization and cultural adaptation efforts by reducing the scale and resources required to acquire such data. 
Just as traditional data augmentation methods improve model robustness without the need for entirely new data creation, our proposed approach similarly complements translation by enabling the creation of large-scale, culturally adapted datasets with minimal overhead.\vspace{1mm}

\noindent \textbf{Automatic Entity based Augmentation.} 
\looseness-1 In addition to manual localization efforts carried out by volunteers,~\citet{ye2024llm} introduce a novel data augmentation technique that leverages large language models (LLMs). Specifically, they apply this method to enhance performance in few-shot Named Entity Recognition (NER) tasks, demonstrating that LLM-driven augmentations can serve as a valuable complement to human-curated resources. 

\noindent\textbf{Mathematics and Cultural Entities.} 
~\citet{karim2025lost,tomar2025mathematics} examined the influence of cultural context on mathematical problems by analyzing the impact of culturally grounded entities such as personal names and food items. However, one critical dimension that remains underexplored is the role of language itself and the biases that can be propagated through translation. Since the significant number of multilingual training and evaluation datasets are created via translation from high-resource languages~\cite{alabi2025charting}, they often fail to capture the linguistic and cultural nuances of the target languages. Evaluating model robustness in the presence of culturally specific entities is a critical challenge, especially as LLMs are increasingly deployed in real-world applications. Native language speakers naturally refer to familiar names, organizations, and currencies from their own cultural context, making it essential for models to handle such variations reliably.

In this work, we advance multilingual evaluation by establishing cultural robustness as a first-class axis of analysis, systematically demonstrating how language choice and culturally specific entities impact model performance and expose biases arising from English-centric entities commonly preserved in translated benchmarks. We introduce an automated and controlled localization pipeline that operationalizes entity replacement at scale across 18 low-resource languages, explicitly decouples translation from cultural localization, and enforces structural and semantic consistency through automated quality-control mechanisms. Building on this framework, we release large-scale culturally localized benchmark variants that enable principled robustness evaluation and effective fine-tuning, supporting both scalable generalization and reliable performance under culturally grounded conditions.


\section{Methodology}

\subsection{Pipeline}
\label{sec:pipeline}

Figure~\ref{stages}, shows our proposed automated pipeline for socio-cultural localization on MWPs. We first extract personal names, organization names, and currencies, and then replace them with manually collected local entities. Then, we generate accurately localized training and test sets for our experiments. Unlike LLM-based localization, our approach ensures that all relevant entities, particularly those critical to the problem, are consistently and correctly replaced. Additionally, we incorporate manual verification steps to further improve localization quality and ensure high fidelity in culturally adapting the benchmarks.

Below, we outline the key stages of our framework pipeline and discuss the design choices that guided its development:

\noindent\textbf{Entity Classification.} Our pipeline processes each word in the input text and classifies it as a personal name, organization name, or currency. We chose to focus on these entity types to enable controlled generation and replacement, ensuring that the original meaning of the sentence is preserved and that no unintended or confusing content is introduced. Replacing animal and food names tend to generate sentences that lack contextual meaning even though our pipeline can handle it easily. While we evaluated several Named Entity Recognition (NER) methods~\cite{tjong-kim-sang-de-meulder-2003-introduction} and POS tagging using spaCy\footnote{\url{https://spacy.io/usage/linguistic-features}}, we found that large language models (LLMs) with structured output formats were significantly more robust. In particular, they handled variations in spelling and casing more effectively, which are common failure points for traditional models. Additionally, using LLMs for entity classification enables a more scalable pipeline, allowing for easy extension to include additional entity types based on the requirements of the task. 

\noindent\textbf{Multilingual Entity Database.} To ensure that the entities used for replacement were culturally relevant, we collaborated with a team of volunteers to curate unisex personal names, organization names, and representative currency values for each language. Special attention was given to selecting unisex names to avoid introducing gender-specific biases or incorrect pronoun associations.\vspace{1.5mm}

\noindent\textbf{Replacement Dictionary Creation.}
Using the output of the entity classification step, we assign replacement entities to each extracted item by referencing a multilingual entity database, as illustrated in Stage 4 of Table~\ref{stages}.\vspace{1.5mm}

\noindent\textbf{Auto Localization.}
Our pipeline operates on three versions of each input sample: the original English sentence (\( x_{\text{en}} \)), its direct translation into a target language (\( x_{\text{trans}} \)), and an entity-replaced version of the translation (\( x_{\text{ent}} \)), as illustrated in Table~\ref{stages}. 

To generate a properly localized version of \( x_{\text{ent}} \), we use a one-shot prompting setup. Specifically, we construct a prompt by showing the LLM the pair \( (x_{\text{en}}, x_{\text{trans}}) \) as an example, and then ask it to translate \( x_{\text{ent}} \) accordingly:

\[
\text{LLM} \left( \left[ (x_{\text{en}}, x_{\text{trans}}),\ x_{\text{ent}} \right] \right) \rightarrow \hat{x}_{\text{loc}}
\]

This method provides stronger contextual grounding and helps the model preserve the structure and fluency of the target language. Empirically, we found it to be more effective than directly prompting the LLM to localize text without reference examples.\vspace{1.5mm}

\noindent\textbf{Quality Check Blocks.}
Between most of the modules in our pipeline, we applied lightweight quality checks to ensure accurate and consistent localization. The core motivation for building this controlled pipeline, as opposed to relying solely on fully automated localization with LLMs, is to guarantee that the model either produces a localized version of the text or returns the original non-localized text if no entities are detected. This conditional behavior prevents unnecessary modifications and maintains data integrity.


As shown in stage 7 of Table \ref{stages}, our quality control measures ensure no overlapping or inconsistent entity replacements, enforce a single currency type per problem to avoid conversion errors, verify that localized outputs match the length of their direct translations, and check similarity between localized and translated text using the difflib library.\footnote{\url{docs.python.org/3/library/difflib.html}} Since all languages share the same entity-replaced text ($x_{\text{ent}}$), human verification only requires comparing entities in the replacement dictionary. This prevents unnecessary additions by LLMs, maintains prompt consistency, and keeps outputs close to the original structure.

\looseness-1\noindent\textbf{LLMs for Localization.}
In this work, we leveraged Gemini models~\cite{comanici2025gemini} due to their exceptional multilingual performance shown by natural language generation (NLG) multilingual benchmarks~\cite{ojo2023good}. To reduce experimental costs while assessing our approach, we used \geminionepro~for evaluation of the framework, and employed \geminitwopro~for the final data generation.

\subsection{Datasets Used}



\looseness-1\noindent\textbf{Evaluation Dataset.}
AfriMGSM~\cite{adelani2024irokobench} is a manually translated benchmark spanning 18 languages sourced from MGSM~\cite{shi2022language}. We found that $\approx$86\% of the test set, includes at least one important entity. Using the English and manually translated pairs, we created a localized evaluation dataset by replacing these entities with culturally appropriate local alternatives while keeping the rest of the problem content unchanged. The correctness of the localized dataset was verified through manual inspection by the authors. Since the approach returns the unlocalized item if it fails, no additional noise is introduced.

\begin{table}[!t]
    \centering
    \label{datasplit}
    \resizebox{\linewidth}{!}{
    \begin{tabular}{l|l|l|l|l}
    \toprule
        \textbf{source } & \textbf{split}  & \textbf{\#source}& \textbf{\#localized} & \textbf{\#lang.} \\ \midrule
        GSM8K & train & 8790 & 25100 & 18 \\ \hline
        Localized-AfriMGSM & test & 4500 & 4500 & 18 \\ \bottomrule
    \end{tabular}
    }
    \caption{Datasets used in our auto localization framework and their details (sources, size, split).}\vspace{-2mm}
    
\end{table}

\noindent\textbf{Training Dataset.}
Due to the lack of manually translated training datasets for math word problems across all target languages, we leveraged the GSM8K dataset from~\citet{cobbe2021training} and applied our pipeline to generate translated and localized versions. For translation, we used the NLLB-200-3.3B model~\cite{nllb2022}, which we consider to offer the best trade-off between model size and translation quality among open-source models that support all the languages considered in this work. 

To ensure translation quality, we filtered the outputs using SSA-COMET~\cite{li2025ssa}, a sentence-level semantic similarity metric that scores translation quality on a scale from 0 to 1, with 1 indicating perfect translation. Based on the score distribution shown in the Appendix \ref{sec:commetscore}, we retained only translations with a COMET score above 0.65. From these, we selected the top 1,500 examples per language for automatic localization and use as training data in our after localizing each. While intermediate experiments leverage \geminionepro~to evaluate the pipeline and assess its effectiveness due to cost constraints and to avoid unwanted behaviors, such as prompt caching\footnote{\url{https://ai.google.dev/gemini-api/docs/caching?lang=python}}, the final dataset is generated using \geminitwopro~to ensure the highest quality standards.

\begin{figure}[t]
\includegraphics[width=0.49\textwidth]{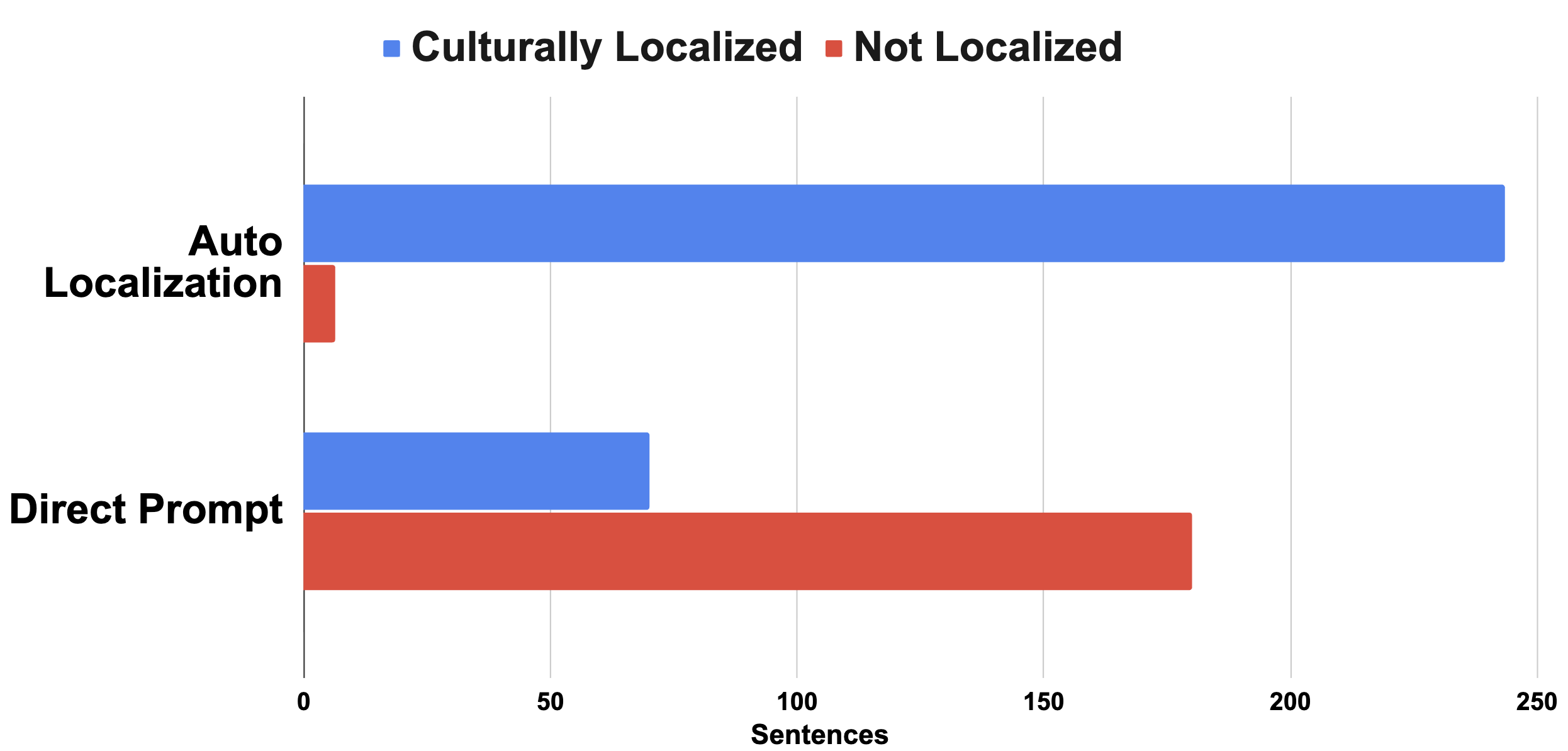}\vspace{-3mm}
\caption{\textbf{Human Validated Comparison of Localization Quality Between Auto Localization and Direct Prompting }(\geminionepro). Our auto localization framework produces significantly better and appropriate culturally localized outputs compared to direct prompting, which often fails to adapt entities to the target culture. 
}\label{localization-stat}

\end{figure}

\begin{figure*}[!ht]
\centering
\includegraphics[width=\textwidth]{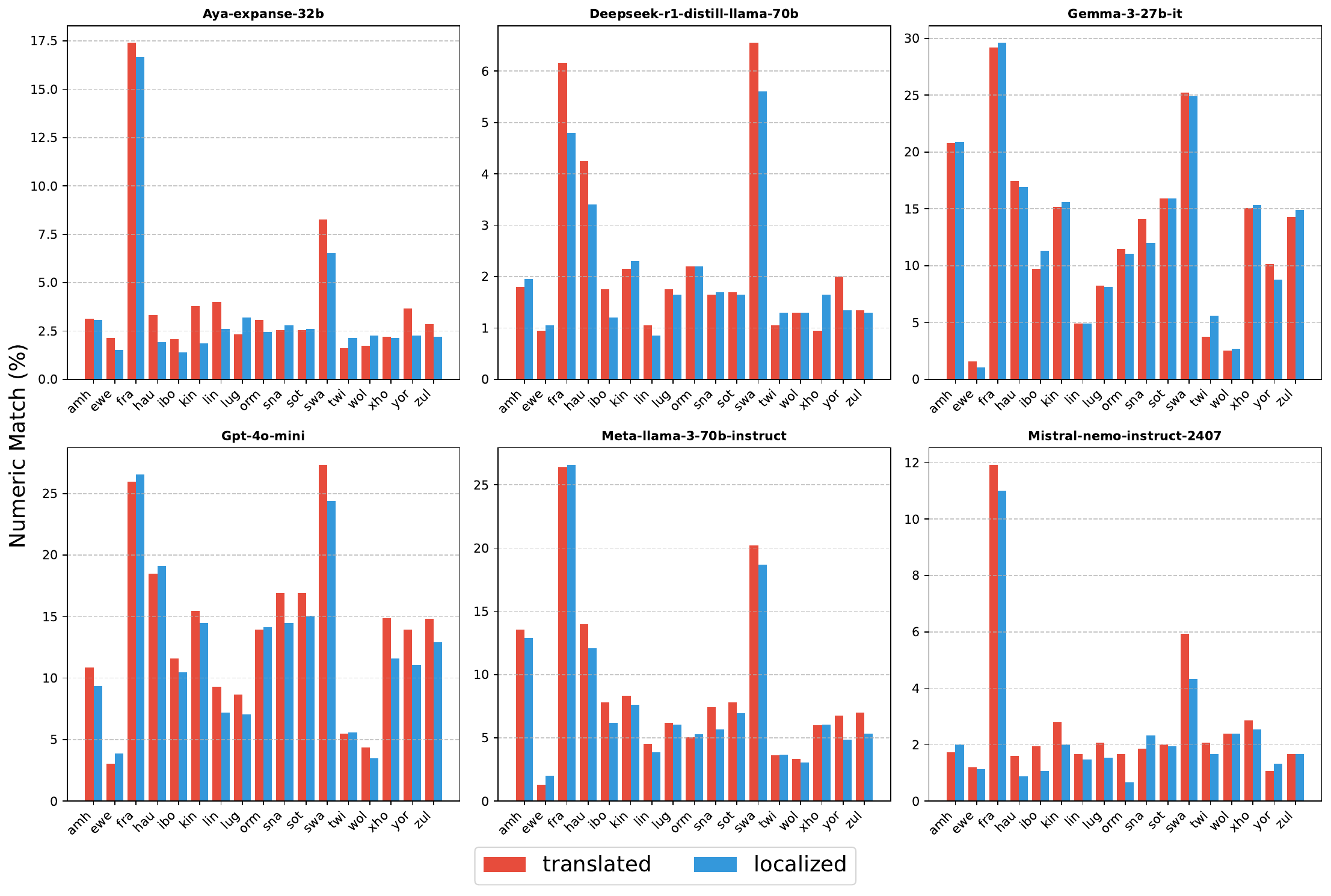}
\vspace{-3mm}
\caption{\textbf{\colorbox{myRed}{Direct Translation}}(\textit{AfriMGSM}) \textbf{vs. \colorbox{cyan}{Auto Localization}}(\textit{Localized-AfriMGSM}) \textbf{Numeric Match performance.} We observe performance differences between translated and localized benchmark indicating a lack of robustness in LLM mathematical ability for real-life culturally localized MWP variants.  }
\label{native_robustness_in_native_language}
\end{figure*}




\begin{figure*}[h]
 \centering
    \includegraphics[width=0.99\textwidth]{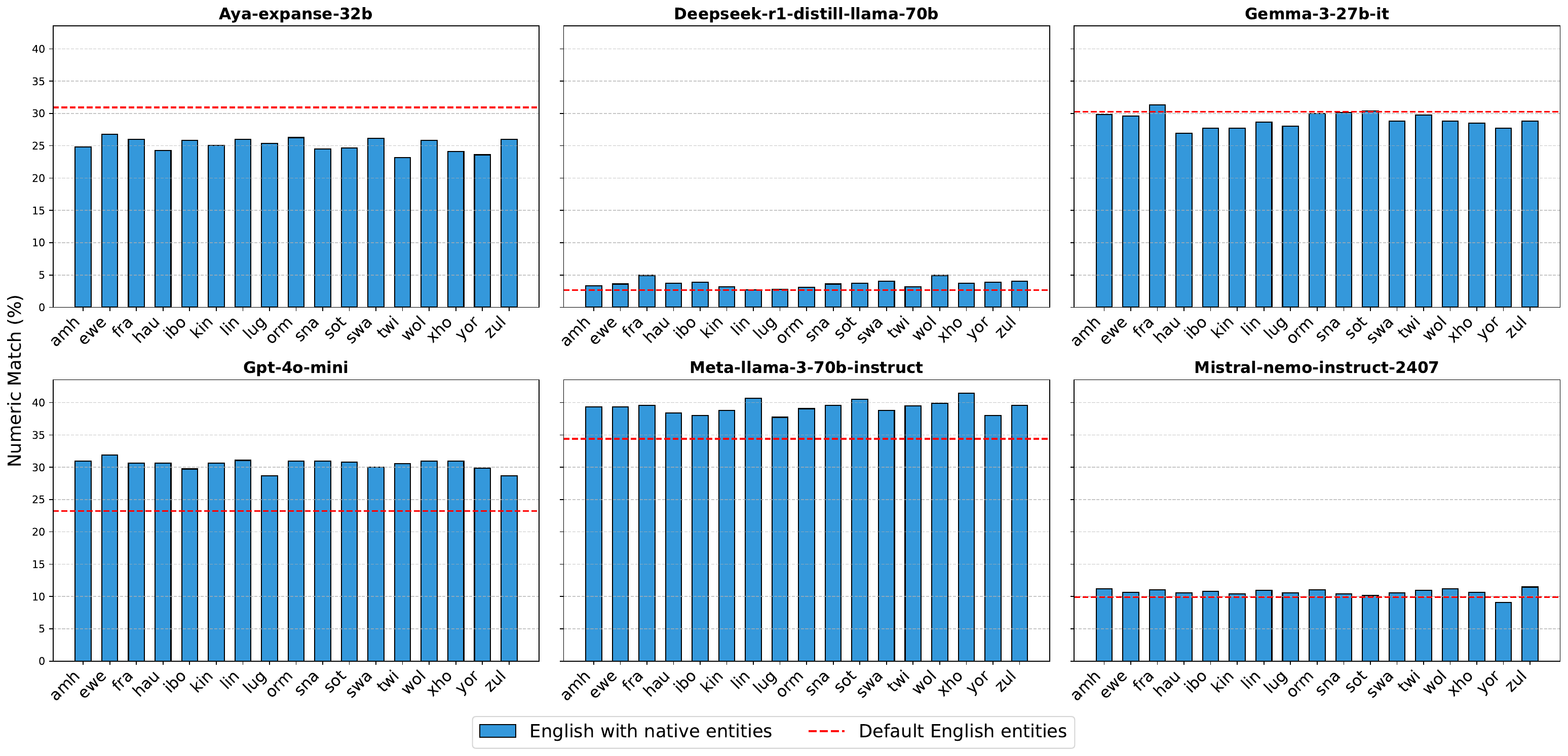}
\caption{\textbf{Effect of Cultural Entities on English Benchmarks.} 
    We investigate whether replacing default English entities with culturally specific entities (\(x_{ent}\); see Table~\ref{stages}) influences model performance. 
    The results show that across models and languages, the inclusion of local entities consistently shifts evaluation outcomes, 
    indicating that cultural grounding plays a measurable role in benchmark performance.}
\label{eng_with_entities}

\end{figure*}

\section{Experimental Setup}

\noindent\textbf{Evaluated LLMs.}
In this work, we evaluated both open-source and closed-source models commonly studied in existing multilingual mathematical research. Our selection spans a range of model sizes, including \cohere, \deepseek, \gemmanineit, \gptfmini, \llamasit ~and \nemo ~\cite{dang2024ayaexpansecombiningresearch},  The finetuning experiments are done only on \llamathree, \gemmanineit~because of compute constraints. \vspace{1.5mm}

\noindent\textbf{Evaluation Metrics.}
In this work we leveraged \textbf{Numeric Match (NM)}, metric that checks whether the predicted numerical value matches the ground truth, ignoring differences in formatting such as units or punctuation after converting them into floating point data and account for errors between them. Moreover, for ensuring robustness and minimal noise via prompt selection, we adopted three prompt variations from~\cite{adelani2024irokobench} for our experiments. We further customized them to return only the final output without intermediate steps.
In low-resource and multilingual scenarios, extracting reliable intermediate reasoning steps is challenging due to language-specific expression, inconsistent formatting, and the absence of standardized reasoning annotations across languages. Final-answer evaluation therefore provides a language-agnostic and comparable signal that isolates whether entity-level cultural changes alter task success, independent of how reasoning is verbalized.

Importantly, our analysis focuses on relative performance shifts between translated and culturally localized variants rather than absolute accuracy. Under this controlled comparison, differences in final-answer accuracy can be attributed to sensitivity to cultural entities, even if the underlying failure manifests as reasoning, arithmetic, or formatting errors. We view finer-grained diagnostic analyses as a valuable direction for future work, particularly as multilingual reasoning benchmarks mature.


\section{Results and Analysis}

We now present the results of our experiments across several LLMs and our generated localized datasets to answer each of our RQs:\vspace{2mm}

\looseness-1\noindent\textbf{ \textit{(RQ1) Can LLM-driven pipelines be used to localize translated benchmarks?}} To address the high cost associated with using human translators for creating localized versions of datasets, we introduce an automatic pipeline that generates culturally adapted versions of mathematical word problems. While this pipeline is not intended to replace human annotators, it serves as a valuable tool for producing augmented datasets that help improve the robustness of LLMs to entity-based variations in problems.

In our first set of experiments (please see Figure \ref{localization-stat}), we used \geminionepro~to generate localized versions of the dataset using both our six-stage localization pipeline and direct prompting via the Gemini API. 
Once both localized dataset variants were generated, three independent annotators evaluated each instance and labeled it as either a valid or invalid localization. Inter-annotator agreement was then measured using Cohen’s kappa, yielding strong agreement for both methods: 0.79 for AutoLocalized and 0.84 for DirectPrompting. These results indicate high annotation consistency and support the reliability of the localization pipeline. Importantly, the approach does not rely on the latest or largest language models; comparable performance is achieved using earlier-generation models such as \geminionepro, demonstrating the robustness and efficiency of the proposed method.

Next, we compare final downstream task performance across directly translated and auto localized versions of MWPs. These results are provided in Figure~\ref{localization-stat} and illustrate the effectiveness of our localization pipeline compared to manual prompting. In addition to improved performance and more accurate localization, our method offers greater control over the types of modifications applied. Manual prompting performs well when there are direct equivalents for names across languages, but it often returns the original translation rather than a properly localized version. This limits its effectiveness in producing culturally adapted outputs. 
In Figure~\ref{localization-stat} items that are not localized include,  names or organization names the pipeline was not able to capture and fallback output because quality checker. 

Finally, we observe that our localization pipeline occasionally confuses organization names with personal names, particularly when organization names contain personal identifiers (e.g., ``Dr.\ Wertz's School''). In such cases, the pipeline may incorrectly substitute the entity, leading to unintended replacements of institutions (e.g., schools) with other types of organizations, thereby distorting the original meaning.

To mitigate this issue, we recommend prioritizing the localization of well-defined entities such as person names and currencies, where ambiguity is minimal. Additionally, this dataset is best utilized as a scalable resource for augmenting training data, rather than as a fully reliable source of semantically precise transformations.

\begin{table*}[!ht]
    \centering
    \scalebox{0.8}{
    \begin{tabular}{l|ccccc|ccccc}
    \toprule
        \multirow{2}{*}{Languages} & \multicolumn{5}{c|}{LLaMA-3-8B-Instruct} & \multicolumn{5}{c}{Gemma-2-9b-it} \\ \cline{2-11}
         & $x_{\text{trans}}$ & $x_{\text{ent}}$ & $\hat{x}_{\text{loc}}$ & all data & \# samples 
         & $x_{\text{trans}}$ & $x_{\text{ent}}$ & $\hat{x}_{\text{loc}}$ & all data & \# samples \\ \midrule
        Hausa & \cellcolor{myRed}{-0.13} & \cellcolor{myRed}{-0.27} & \cellcolor{myRed}{-1.20} & \cellcolor{myGreen}{0.80} & 5k & \cellcolor{myRed}{-1.73} & \cellcolor{myGreen}{0.93} & \cellcolor{myGreen}{1.07} & \cellcolor{myGreen}{0.40} & 10k \\ 
        Swahili & \cellcolor{myGreen}{0.40} & \cellcolor{myGreen}{1.33} & \cellcolor{myRed}{-0.53} & \cellcolor{myGreen}{0.67} & 1k & \cellcolor{myRed}{-2.00} & \cellcolor{myGreen}{1.60} & \cellcolor{myGreen}{1.20} & \cellcolor{myGreen}{1.07} & 5k \\ 
        Ewe & \cellcolor{myRed}{-0.67} & \cellcolor{myGreen}{0.13} & \cellcolor{myGreen}{0.27} & \cellcolor{myGreen}{0.80} & 25k & \cellcolor{myGreen}{0.27} & \cellcolor{myRed}{-0.53} & \cellcolor{myGreen}{0.27} & \cellcolor{myGreen}{1.20} & 25k \\ 
        Twi & \cellcolor{myGreen}{0.67} & \cellcolor{myRed}{-0.27} & \cellcolor{myGreen}{0.53} & \cellcolor{myRed}{-0.93} & 10k & \cellcolor{myRed}{-0.27} & \cellcolor{myGreen}{0.13} & \cellcolor{myRed}{-0.53} & \cellcolor{myGreen}{0.67} & 1k \\ 
        Wolof & \cellcolor{myRed}{-0.27} & \cellcolor{myRed}{-0.13} & \cellcolor{myGreen}{0.93} & \cellcolor{myGreen}{0.93} & 1k & \cellcolor{myRed}{-0.40} & \cellcolor{myRed}{-0.27} & \cellcolor{myGreen}{0.27} & \cellcolor{myGreen}{0.67} & 5k \\ 
        Lingala & \cellcolor{myRed}{-0.13} & \cellcolor{myRed}{-0.67} & \cellcolor{myGreen}{0.27} & \cellcolor{myGreen}{0.67} & 1k & \cellcolor{myGreen}{0.40} & \cellcolor{myGreen}{0.67} & \cellcolor{myGreen}{2.00} & \cellcolor{myGreen}{1.47} & 10k \\ 
        Luganda & \cellcolor{myRed}{-0.13} & \cellcolor{myGreen}{0.13} & \cellcolor{myYellow}{0.00} & \cellcolor{myGreen}{0.40} & 1k & \cellcolor{myGreen}{0.13} & \cellcolor{myRed}{-1.59} & \cellcolor{myGreen}{1.07} & \cellcolor{myGreen}{0.27} & 5k \\ 
        Oromo & \cellcolor{myRed}{-0.40} & \cellcolor{myRed}{-0.80} & \cellcolor{myRed}{-0.53} & \cellcolor{myGreen}{0.67} & 10k & \cellcolor{myRed}{-0.27} & \cellcolor{myRed}{-0.40} & \cellcolor{myGreen}{0.80} & \cellcolor{myRed}{-1.07} & 1k \\ 
        Shona & \cellcolor{myRed}{-1.73} & \cellcolor{myRed}{-0.27} & \cellcolor{myRed}{-0.13} & \cellcolor{myGreen}{0.93} & 1k & \cellcolor{myGreen}{2.27} & \cellcolor{myRed}{-0.27} & \cellcolor{myGreen}{1.20} & \cellcolor{myGreen}{1.60} & 1k \\ 
        Xhosa & \cellcolor{myRed}{-0.27} & \cellcolor{myRed}{-1.20} & \cellcolor{myGreen}{0.27} & \cellcolor{myGreen}{0.93} & 1k & \cellcolor{myGreen}{0.67} & \cellcolor{myYellow}{0.00} & \cellcolor{myGreen}{1.07} & \cellcolor{myGreen}{1.87} & 1k \\ 
        Yoruba & \cellcolor{myRed}{-1.07} & \cellcolor{myRed}{-0.67} & \cellcolor{myRed}{-0.40} & \cellcolor{myGreen}{0.27} & 1k & \cellcolor{myYellow}{0.00} & \cellcolor{myGreen}{0.13} & \cellcolor{myGreen}{1.20} & \cellcolor{myGreen}{0.13} & 1k \\ 
        Kinyarwanda & \cellcolor{myGreen}{0.27} & \cellcolor{myRed}{-0.27} & \cellcolor{myRed}{-0.40} & \cellcolor{myGreen}{1.73} & 1k & \cellcolor{myYellow}{0.00} & \cellcolor{myRed}{-1.60} & \cellcolor{myGreen}{0.67} & \cellcolor{myGreen}{0.67} & 25k \\ 
        Zulu & \cellcolor{myGreen}{0.27} & \cellcolor{myRed}{-1.07} & \cellcolor{myGreen}{0.80} & \cellcolor{myRed}{-0.80} & 1k & \cellcolor{myRed}{-0.67} & \cellcolor{myGreen}{1.20} & \cellcolor{myGreen}{0.40} & \cellcolor{myGreen}{0.13} & 1k \\ 
        Sotho & \cellcolor{myRed}{-0.40} & \cellcolor{myYellow}{0.00} & \cellcolor{myGreen}{0.40} & \cellcolor{myGreen}{0.53} & 1k & \cellcolor{myYellow}{0.00} & \cellcolor{myRed}{-1.60} & \cellcolor{myGreen}{0.67} & \cellcolor{myGreen}{0.67} & 10k \\ 
        Igbo & \cellcolor{myYellow}{0.00} & \cellcolor{myRed}{-0.27} & \cellcolor{myGreen}{0.13} & \cellcolor{myGreen}{0.67} & 1k & \cellcolor{myRed}{-0.53} & \cellcolor{myGreen}{1.87} & \cellcolor{myGreen}{0.67} & \cellcolor{myGreen}{0.27} & 1k \\ 
        Amharic & \cellcolor{myGreen}{0.13} & \cellcolor{myRed}{-1.20} & \cellcolor{myGreen}{1.07} & \cellcolor{myGreen}{0.93} & 1k & \cellcolor{myRed}{-1.73} & \cellcolor{myGreen}{0.93} & \cellcolor{myGreen}{1.07} & \cellcolor{myGreen}{0.40} & 25k \\ 
        French & \cellcolor{myGreen}{1.07} & \cellcolor{myGreen}{0.93} & \cellcolor{myGreen}{0.67} & \cellcolor{myRed}{-0.93} & 5k & \cellcolor{myRed}{-1.07} & \cellcolor{myRed}{-1.33} & \cellcolor{myRed}{-0.93} & \cellcolor{myGreen}{0.53} & 10k \\ \midrule
        \# Native Robust Lang. & 6 & 4 & 10 & 14 & ~ & 5 & 8 & 15 & 16 & ~ \\ \bottomrule
    \end{tabular}
    }
    \caption{\textbf{Native Robustness (Numeric Match $\Delta$).}
We report $\Delta_{\mathrm{NM}}=\mathrm{NM}_{\text{localized}}-\mathrm{NM}_{\text{translated}}$ across sampled data fine-tunings for translated data ($x_{\text{trans}}$), English entity–replaced data ($x_{\text{ent}}$), auto-localized data ($\hat{x}_{\text{loc}}$), and all data combined. \colorbox{myGreen}{Positive} values indicate higher robustness on localized benchmarks, \colorbox{myRed}{Negative} values indicate stronger performance on English-centric benchmarks, and \colorbox{myYellow}{Yellow} denotes no change.}
\label{performance}
\vspace{-3mm}
\end{table*}

\vspace{1.5mm}
\noindent\textbf{ \textit{(RQ2) Does introducing local entities in translated benchmarks reveal performance disparities for MWPs in LRLs and English?}} We first analyze LRLs and then English language MWPs. Given that 30\% of newly created African datasets are based on translated content~\cite{alabi2025charting}, it is important to assess whether LLMs are overfitting to English-centric entities commonly preserved during translation. Such overfitting may lead to performance degradation when these entities are replaced with their native counterparts. In this work, we hypothesize that models should perform better when evaluated on data containing english entities because english centric bias that is found in most training and evaluation dataset. 

Figure~\ref{native_robustness_in_native_language} illustrates the performance differences between models evaluated on translated benchmarks and those evaluated on automatically localized benchmarks, highlighting the impact of entity localization on model robustness. We can observe that the all models except \gemmatwentyit~perform well on translated benchmarks. This shows the translated benchmarks tend to mislead the performance of models and the evaluation doesn't directly simulate the real life usage when people use their local entities in the problems.

\gemmatsit~demonstrates more consistent results across both native and English-centric benchmarks. At the language-specific level, we observe that French and Swahili, relatively higher-resource languages, show pronounced effects in the \deepseek~and \nemo ~models. 

Next, Figure~\ref{eng_with_entities} shows the performance of math word problems in English where personal names, organization names, and currencies have been replaced with entities from the respective target languages. We compare these results with baseline scores obtained from the original English benchmark, which contains English-centric names and currencies. This comparison allows us to evaluate the impact of introducing culturally specific entities into English problem statements.

Both \llamasit~and \gptfmini~achieve higher accuracy on the localized (native-entity) variant than on the English benchmark, whereas the \cohere~model’s performance lags behind. \gemmatsit~generally tracks the localized variant more closely, despite occasional dips and overshoots. We have relatively similar trends across languages unlike LRL related experiments.

\paragraph{ \textit{(RQ3) Can automatic localization improve model robustness by augmenting benchmarks with culturally adapted variants?}} Figure~\ref{performance} presents the performance of models  on different flavors of MWP data. 
From the multilingual dataset we created through translation ($x_{\text{trans}}$), English entity–replacement ($x_{\text{ent}}$), auto-localization ($\hat{x}_{\text{loc}}$), and a combination of all datasets, we randomly sampled subsets of 1k, 5k, 10k, and the full 25k examples for model training. We opted for various training set sizes since different languages require different volumes of training data to enhance model robustness across localized MWP versions. 

Evaluation was then conducted on both the original translated datasets and our localized versions to assess performance differences achieved. This comparison helps us understand whether the inclusion of native cultural entities affects model behavior. 
We observe that both \llamathreeit~and \gemmanineit~models exhibit improved robustness to entity changes in several languages. For both models, the best performance is achieved when localization is combined with additional noisy data, indicating that diverse training sources can enhance generalization. 

Looking at the translated ($x_{\text{trans}}$) and English entity–replaced ($x_{\text{ent}}$) data, \gemmanineit~demonstrates stronger performance when local entities are present in the questions, whereas \llamathreeit~exhibits a performance drop. Incorporating localized datasets in addition to language changes in the training set leads to improvements over purely translated benchmarks in both models, though \gemmanineit~benefits more from this effect. Finally, combining all datasets yields models that are more robust to these variations.


\section{Conclusion}
Due to the scarcity of native, low-resource mathematical reasoning datasets that include local entities, translation remains the predominant source of benchmark questions. However, performance on these translated benchmarks is highly sensitive to English-centric terms. We present a framework that culturally localizes translated datasets into variants enriched with local entities . We highlight the biases and instabilities introduced by translation-only benchmarks and show that our localization framework improves model robustness in the presence of native entities.

Greater emphasis should be placed on creating MWPs that center around the cultural activities of the target community, in addition to incorporating cultural names into existing ones. For future work, we will use this framework for more reasoning-focused evaluations, create strong multilingual training datasets, and extend the approach beyond mathematics to complement translation-based benchmarking in other tasks.

\section*{Limitations}

Due to the high cost associated with human-centered localization, we developed an automatic localization pipeline capable of generating culturally relevant datasets from translated, English-centric data. While this pipeline offers a scalable and efficient solution for large-scale dataset creation, it is not intended to replace human annotation. In scenarios where the cost of data acquisition is not a constraint, human-centered localization should be preferred for its higher accuracy and cultural fidelity. The primary advantage of our automated approach lies in its ability to support the expansion of training pipelines across multiple languages and domains with minimal manual effort. We galvanize the community to place greater emphasis on developing math word problems rooted in everyday community activities, creating appropriate socio-cultural scenarios that can be used to further improve models.

\section*{Acknowledgment}

The authors would like to thank the German Federal Ministry of Education and Research and the German federal states (http://www.nhr-verein.de/en/our-partners) for supporting this work/project as part of the National High-Performance Computing (NHR) joint funding program. We also thank PaliGemma academic program for gemini credits used in this research.

\bibliography{custom}

\appendix

\section*{Appendix}
\section{Full Sampling result}

As shown in Table~\ref{full:sample}, the results provide a comprehensive view of how different sampling sizes and training configurations affect robustness across languages. Several key observations emerge. First, the effect of data localization is not uniform across models: while LLaMA-3-8B-Instruct often exhibits fluctuations depending on sample size, Gemma-2-9b-it tends to benefit more consistently from auto-localized data ($\hat{x}_{\text{loc}}$). This suggests that Gemma is more sensitive to entity grounding and gains robustness when trained with culturally and linguistically aligned examples.

\begin{table}[!ht]
    \centering
    \scalebox{0.5}{
    \begin{tabular}{l|llll|llll|l}
    \toprule
        \multirow{2}{*}{Lang}  & \multicolumn{4}{c|}{LLaMA-3-8B-Instruct} &\multicolumn{4}{c}{Gemma-2-9b-it} & \textbf{} \\ \cline{2-10}
        & $x_{\text{trans}}$ & $x_{\text{ent}}$ & $\hat{x}_{\text{loc}}$ & all data & $x_{\text{trans}}$ & $x_{\text{ent}}$ & $\hat{x}_{\text{loc}}$ & all data & \# sample \\ \midrule
        amh & 0.13 & -1.20 & 1.07 & 0.93 & 0.53 & -0.53 & -0.93 & 0.80 & 1000 \\ \
        amh & -0.27 & -0.13 & 1.33 & 0.40 & 0.27 & 0.00 & 0.40 & -1.20 & 5000 \\ \
        amh & 1.33 & -0.27 & 0.53 & -1.07 & -0.13 & 0.40 & -0.53 & -0.13 & 10000 \\ \
        amh & 1.33 & -0.27 & 0.53 & -1.07 & -1.73 & 0.93 & 1.07 & 0.40 & 25100 \\ \
        ewe & -0.67 & 0.13 & 0.27 & 0.80 & 0.27 & -0.53 & 0.27 & 1.20 & 25100 \\ \
        ewe & -0.67 & 0.13 & 0.27 & 0.80 & 0.27 & -0.53 & 0.27 & 1.20 & 10000 \\ \
        ewe & 0.40 & 0.27 & -0.13 & -0.67 & 0.13 & 0.13 & 0.27 & -0.67 & 1000 \\ \
        ewe & -0.27 & -0.40 & 0.13 & -1.33 & 0.53 & -0.93 & 1.87 & -0.67 & 5000 \\ \
        fra & 1.07 & 0.93 & 0.67 & -0.93 & 0.40 & -0.53 & -0.27 & -0.13 & 5000 \\ \
        fra & -0.53 & 0.67 & -0.67 & -0.67 & -1.07 & -1.33 & -0.93 & 0.53 & 10000 \\ \
        fra & -1.60 & -0.53 & -0.67 & -0.67 & -0.67 & 2.40 & 0.80 & -0.93 & 1000 \\ \
        fra & -0.53 & 0.67 & -0.67 & -0.67 & -0.13 & 0.40 & -0.53 & -0.13 & 25100 \\ \
        hau & -0.13 & -0.27 & -1.20 & 0.80 & -2.13 & -1.07 & -1.87 & 0.00 & 5000 \\ \
        hau & -0.80 & -0.80 & -1.20 & -0.27 & -1.73 & 0.93 & 1.07 & 0.40 & 10000 \\ \
        hau & -0.93 & -0.27 & -1.60 & -0.93 & -0.40 & -1.33 & -0.67 & -2.53 & 1000 \\ \
        hau & -0.80 & -0.80 & -1.20 & -0.27 & -1.20 & -0.93 & 0.67 & -0.93 & 25100 \\ \
        ibo & 0.00 & -0.27 & 0.13 & 0.67 & -0.53 & 1.87 & 0.67 & 0.27 & 1000 \\ \
        ibo & 0.67 & 0.13 & 0.80 & -0.80 & -0.67 & 0.67 & 0.67 & 0.13 & 5000 \\ \
        ibo & 0.53 & 0.80 & 0.67 & -0.93 & -0.93 & 0.67 & 0.27 & -1.33 & 10000 \\ \
        ibo & 0.53 & 0.80 & 0.67 & -0.93 & -0.40 & -0.13 & -0.40 & -1.07 & 25100 \\ \
        kin & 0.27 & -0.27 & -0.40 & 1.73 & 0.80 & -0.13 & 1.33 & 3.60 & 1000 \\ \
        kin & -1.07 & -0.67 & 0.40 & -0.27 & 0.00 & -1.60 & 0.67 & 0.67 & 25100 \\ \
        kin & 0.00 & 1.20 & -0.53 & 0.53 & -0.27 & 1.07 & -1.73 & -1.47 & 5000 \\ \
        kin & -1.07 & -0.67 & 0.40 & -0.27 & 0.00 & -0.13 & -0.53 & 0.27 & 10000 \\ \
        lin & -0.13 & -0.67 & 0.27 & 0.67 & -0.27 & -0.13 & -1.20 & 0.67 & 1000 \\ \
        lin & -1.07 & -0.80 & -0.13 & -0.67 & 0.40 & 0.67 & 2.00 & 1.47 & 10000 \\ \
        lin & -1.07 & -0.80 & -0.13 & -0.67 & 0.40 & 0.67 & 2.00 & 1.47 & 25100 \\ \
        lin & -0.27 & -0.40 & -0.67 & 0.40 & -1.60 & 0.53 & 0.80 & 0.13 & 5000 \\ \
        lug & -0.13 & 0.13 & 0.00 & 0.40 & -1.07 & -1.47 & -0.13 & -0.13 & 1000 \\ \
        lug & -0.80 & -1.07 & 0.00 & -1.07 & 0.13 & -1.59 & 1.07 & 0.27 & 5000 \\ \
        lug & -1.33 & -0.13 & -0.27 & -1.33 & -2.40 & -0.26 & 0.93 & -1.33 & 10000 \\ \
        lug & -1.33 & -0.13 & -0.27 & -1.33 & -2.40 & -0.26 & 0.93 & -1.33 & 25100 \\ \
        orm & -0.40 & -0.80 & -0.53 & 0.67 & 0.40 & -0.40 & -0.13 & -1.07 & 10000 \\ \
        orm & -0.40 & -0.80 & -0.53 & 0.67 & 0.40 & -0.40 & -0.13 & -1.07 & 25100 \\ \
        orm & -0.40 & 0.27 & -0.80 & -0.53 & -0.27 & -0.40 & 0.80 & -1.07 & 1000 \\ \
        orm & -0.93 & -0.53 & -1.33 & -0.27 & 0.27 & 0.80 & -0.27 & -1.07 & 5000 \\ \
        sna & -1.73 & -0.27 & -0.13 & 0.93 & 2.27 & -0.27 & 1.20 & 1.60 & 1000 \\ \
        sna & 0.00 & 1.07 & 0.13 & -0.80 & 0.80 & 0.13 & 1.87 & -0.40 & 5000 \\ \
        sna & -2.00 & -0.53 & -0.53 & 0.13 & -0.13 & 0.53 & -0.40 & 0.27 & 10000 \\ \
        sna & -2.00 & -0.53 & -0.53 & 0.13 & 0.00 & -0.13 & -0.53 & 0.27 & 25100 \\ \
        sot & -0.40 & 0.00 & 0.40 & 0.53 & 1.07 & 0.80 & 2.93 & -0.13 & 1000 \\ \
        sot & 0.40 & 0.80 & -1.33 & 0.27 & 0.00 & -1.60 & 0.67 & 0.67 & 10000 \\ \
        sot & 0.00 & 0.67 & -0.80 & 1.60 & 0.67 & -1.07 & 0.00 & -1.33 & 5000 \\ \
        sot & 0.40 & 0.80 & -1.33 & 0.27 & -1.07 & -1.33 & -0.93 & 0.53 & 25100 \\ \
        swa & 0.40 & 1.33 & -0.53 & 0.67 & -0.40 & 0.00 & 0.27 & -0.13 & 1000 \\ \
        swa & -0.67 & -1.60 & -0.13 & 0.67 & -2.00 & 1.60 & 1.20 & 1.07 & 5000 \\ \
        swa & -0.67 & -1.20 & 0.40 & 0.27 & -1.07 & -1.33 & 0.40 & -0.40 & 10000 \\ \
        swa & -0.67 & -1.20 & 0.40 & 0.27 & -1.07 & -1.33 & 0.40 & -0.40 & 25100 \\ \
        twi & 0.67 & -0.27 & 0.53 & -0.93 & 2.00 & 0.53 & 1.47 & -0.40 & 10000 \\ \
        twi & 0.67 & -0.27 & 0.53 & -0.93 & 2.00 & 0.53 & 1.47 & -0.40 & 25100 \\ \
        twi & 0.53 & -1.07 & -0.13 & -0.40 & -0.27 & 0.13 & -0.53 & 0.67 & 1000 \\ \
        twi & 1.20 & 0.80 & 0.40 & -0.13 & -1.47 & -0.80 & -0.80 & -0.13 & 5000 \\ \
        wol & -0.27 & -0.13 & 0.93 & 0.93 & 0.13 & 0.80 & 0.40 & 0.27 & 1000 \\ \
        wol & -0.53 & -0.40 & -0.40 & -0.40 & -0.40 & -0.27 & 0.27 & 0.67 & 5000 \\ \
        wol & 0.40 & 0.67 & -0.67 & -0.53 & 0.93 & -0.40 & 0.93 & 0.00 & 10000 \\ \
        wol & 0.40 & 0.67 & -0.67 & -0.53 & 0.93 & -0.40 & 0.93 & 0.00 & 25100 \\ \
        xho & -0.27 & -1.20 & 0.27 & 0.93 & 0.67 & 0.00 & 1.07 & 1.87 & 1000 \\ \
        xho & 0.13 & -0.93 & -0.40 & -0.53 & -1.73 & 0.27 & -1.60 & 0.40 & 5000 \\ \
        xho & -0.67 & -0.27 & -0.13 & -0.67 & -0.40 & -0.13 & -0.40 & -1.07 & 10000 \\ \
        xho & -0.67 & -0.27 & -0.13 & -0.67 & -0.53 & 0.40 & 0.00 & -1.33 & 25100 \\ \
        yor & -1.07 & -0.67 & -0.40 & 0.27 & 0.00 & 0.13 & 1.20 & 0.13 & 1000 \\ \
        yor & 0.27 & 1.33 & 0.00 & -0.13 & 0.80 & -0.27 & 1.07 & 0.00 & 5000 \\ \
        yor & -0.93 & 0.27 & -0.27 & -0.93 & -1.20 & -0.93 & 0.67 & -0.93 & 10000 \\ \
        yor & -0.93 & 0.27 & -0.27 & -0.93 & -0.13 & 0.53 & -0.40 & 0.27 & 25100 \\ \
        zul & 0.27 & -1.07 & 0.80 & -0.80 & -0.67 & 1.20 & 0.40 & 0.13 & 1000 \\ \
        zul & 0.80 & -0.67 & -0.27 & -0.80 & 0.67 & 0.67 & 0.27 & -1.07 & 5000 \\ \
        zul & -1.07 & 0.27 & -0.93 & -1.20 & -0.53 & 0.40 & 0.00 & -1.33 & 10000 \\ \
        zul & -1.07 & 0.27 & -0.93 & -1.20 & -0.93 & 0.67 & 0.27 & -1.33 & 25100 \\ \bottomrule
    \end{tabular}}
    \caption{\textbf{Full Native Robustness (Numeric Match $\Delta$).} 
We report $\Delta_{\mathrm{NM}}=\mathrm{NM}_{\text{localized}}-\mathrm{NM}_{\text{translated}}$ across all sampled data fine-tunings for translated data ($x_{\text{trans}}$), English entity–replaced data ($x_{\text{ent}}$), auto-localized data ($\hat{x}_{\text{loc}}$), and all data combined. 
Positive values indicate higher robustness on localized benchmarks, negative values indicate stronger performance on English-centric benchmarks, and zero denotes no change.}
\label{full:sample}
\end{table}

Second, sample size plays a crucial role. For low-resource settings (e.g., 1k samples), both models display instability, with performance swings that highlight the difficulty of robustly adapting to local contexts with very limited data. As the number of samples increases (e.g., 25k), more stable trends appear, although the direction of improvement still varies by language. For instance, some languages (such as Kinyarwanda and Shona) show strong positive gains under localization, whereas others (such as Hausa and Luganda) exhibit persistent negative or mixed trends, indicating that language-specific features or annotation artifacts may influence results. 

The “all data” setting, combining translated, entity-replaced, and localized data, yields more balanced robustness across languages. This indicates that hybrid augmentation can reduce model brittleness, though closing language gaps requires attention to data quality and localization type, not just scale.

These results underscore the interplay between training data, sampling size, and language characteristics, emphasizing the need to evaluate models on culturally grounded datasets rather than translations alone.
\section{Adopted Evaluation Prompts}

We adapted three prompts from~\cite{adelani2024irokobench} and customized them to ensure that they return only numeric answers.

\begin{tcolorbox}[title=Prompt: Adopted prompt for evaluations, colback=gray!5, colframe=black!70]

\textbf{prompt\_1:} \\
"Question: \{\{lang\}\} \newline
Return the number answer only. Do not provide an explanation. \newline
Number Answer:"

\textbf{prompt\_2:} \\
"Give direct numerical answers for the question provided. \newline
Question: \{\{lang\}\} \newline
Do not provide an explanation. \newline
Numeric Answer:"

\textbf{prompt\_3:} \\
"Solve the following math question. \newline
Question: \{\{lang\}\} \newline
Do not provide an explanation. \newline
Numeric Answer:"

\end{tcolorbox}

\section{Manual Annotation}

For the human annotation shown in Table~\ref{localization-stat}, annotators were asked to examine $x_{\text{ent}}$ and determine whether English-centric names, currencies, or organization names had been replaced. If such entities were replaced, the sample was labeled as \textit{Culturally Localized}; if they remained, it was labeled as \textit{Not Localized}. Since all languages share the same $x_{\text{ent}}$, the manual annotation was carried out once and applied across all languages.

\section{Prompt Templates for the Localization Pipeline}

Below, we present the prompts used for our Auto-Localizer and direct prompt localization. We observed that including the intermediate steps shown in Table~\ref{stages} improves the accuracy of localization.

\begin{tcolorbox}[
    enhanced,
    colback=gray!5!white,
    colframe=black!55,
    title=\textbf{Prompt 3},
    fonttitle=\bfseries,
    arc=2mm,
    boxrule=0.5pt,
    left=2mm,
    right=2mm,
    top=2mm,
    bottom=2mm,
    breakable
]

\textbf{You are an expert linguistic assistant.} Your task is to edit a sentence in a native language to match a change made in its English parallel.

\bigskip

\textbf{Here is the context:}
\begin{itemize}
    \item \textbf{Original English:} The original sentence.
    \item \textbf{Original Native:} The original translation of the English sentence in \texttt{\{native\_lang\}}.
    \item \textbf{Modified English:} The English sentence has been edited. One or more words have been replaced.
\end{itemize}

\textbf{Your goal} is to produce a \textbf{Modified Native} sentence by applying the \emph{exact same replacement} to the \textbf{Original Native} sentence.

\bigskip

\textbf{Crucial Instructions:}
\begin{itemize}
    \item \textbf{DO NOT} re-translate the entire sentence. Only replace the specific words that were changed in the English version.
    \item Preserve the original grammar and structure of the native sentence as much as possible.
    \item Ensure the final \textbf{Modified Native} sentence is natural and grammatically correct in \texttt{\{native\_lang\}}.
    \item Respond with \textbf{ONLY} the \textbf{Modified Native} sentence and nothing else.
\end{itemize}

\bigskip

\textbf{Example:}
\begin{itemize}
    \item Original English: Janet's ducks lay 16 eggs per day.
    \item Original Native (French): Les canards de Janet pondent 16 œufs par jour.
    \item Modified English: Andrea's ducks lay 16 eggs per day.
    \item Modified Native (French): Les canards d'Andrea pondent 16 œufs par jour.
\end{itemize}

\bigskip

\textbf{Your Task:}
\begin{itemize}
    \item Original English: \texttt{\{original\_eng\}}
    \item Original Native (\texttt{\{native\_lang\}}): \texttt{\{original\_native\}}
    \item Modified English: \texttt{\{modified\_eng\}}
    \item Modified Native (\texttt{\{native\_lang\}}): 
\end{itemize}

\end{tcolorbox}

\begin{tcolorbox}[
    enhanced,
    colback=gray!5!white,
    colframe=black!55,
    title=\textbf{Prompt 3},
    fonttitle=\bfseries,
    arc=2mm,
    boxrule=0.5pt,
    left=2mm,
    right=2mm,
    top=2mm,
    bottom=2mm,
    breakable
]

\textbf{You are an expert linguistic assistant specializing in localization.} Your task is to localize the following English sentence into \texttt{\{target\_lang\_name\}}.

\bigskip
\bigskip
\textbf{Localization} means adapting the text to the target culture and context by:
\begin{itemize}
    \item Identifying and replacing English-specific entities (such as person names, organization names, and currency symbols/names) with culturally appropriate equivalents in \texttt{\{target\_lang\_name\}}.
    \item Ensuring the localized text sounds natural and grammatically correct in \texttt{\{target\_lang\_name\}}.
    \item \textbf{Crucially}, the localized text should be generated directly from the English sentence, without relying on any pre-existing native translation.
\end{itemize}


\textbf{English Sentence to Localize:}

\texttt{\{original\_eng\}}

\textbf{Your Goal:} Produce a \textbf{Localized \texttt{\{target\_lang\_name\}}} sentence.


\textbf{Instructions:}
\begin{itemize}
    \item Identify person names, organization names, and currency symbols/names in the English sentence.
    \item Replace these entities with culturally appropriate \texttt{\{target\_lang\_name\}} alternatives.
    \item If an entity does not have a direct cultural equivalent or is already culturally neutral, it may remain unchanged.
    \item Respond with \textbf{ONLY} the \texttt{Localized \{target\_lang\_name\}} sentence and nothing else.
\end{itemize}


\textbf{Localized \texttt{\{target\_lang\_name\}}:}
\end{tcolorbox}



\section{Presence of Cultural entities in our Training data}
\label{train-data-stats}
\begin{figure*}[h]
\centering
\includegraphics[width=0.8\textwidth]{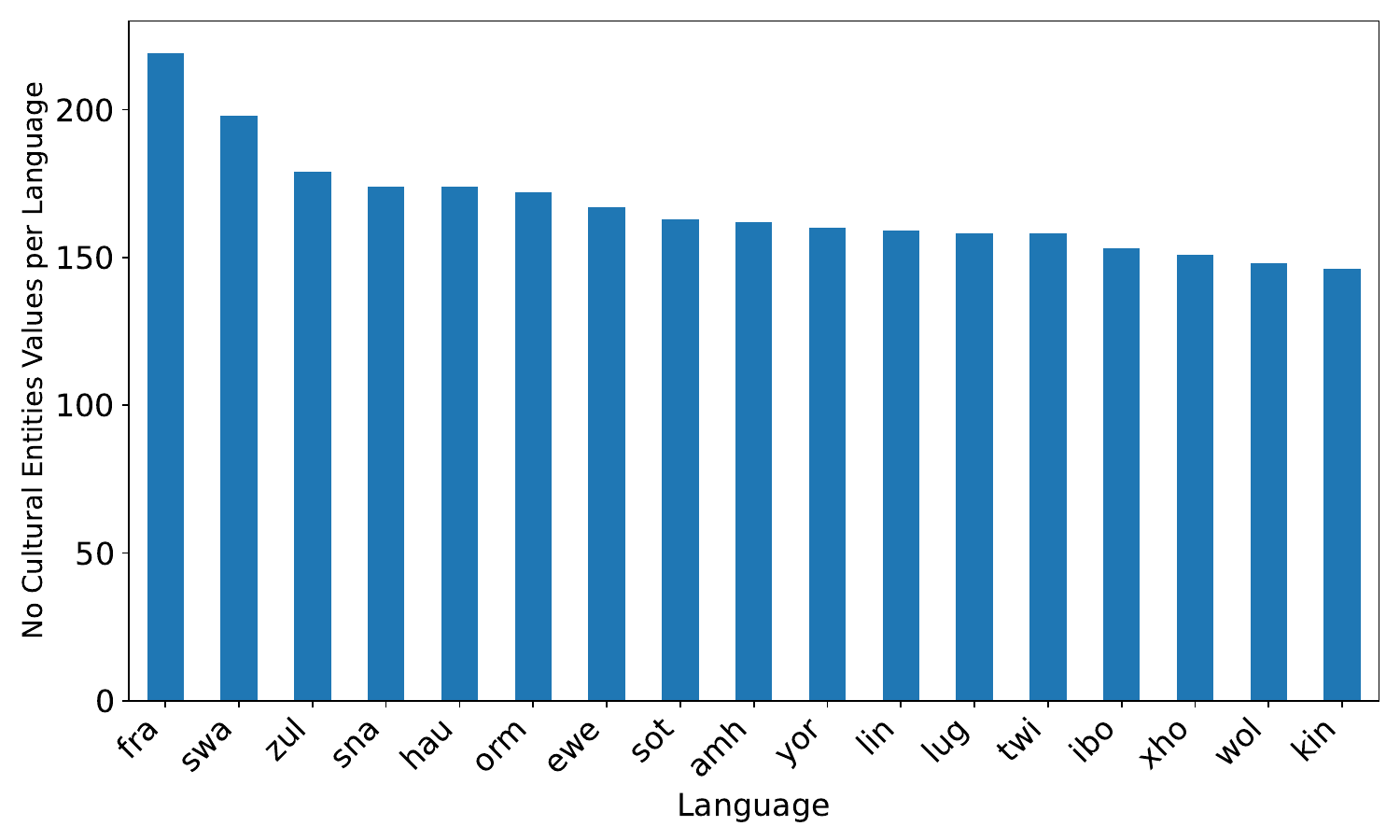}
\caption{Number of data samples without cultural entity replacements (out of 1500 selected), grouped by language in the training dataset.}
\label{training-deepdive}
\end{figure*}

Figure~\ref{training-deepdive} illustrates the distribution of culturally relevant items within the 1,500 data samples selected for each language. Our analysis reveals that a considerable proportion of the translated items could not be localized to the target language, primarily because they did not contain the types of entities, such as personal names, organizations, or currencies, that form the basis of our localization strategy. 

This proportion also helps explain why localized datasets did not improve all languages equally. For some languages, high-quality translations may already omit most culturally salient entities, limiting the added value of localization. In contrast, for languages such as Kinyarwanda, where out of 1500 translations only about 150 lacked cultural references despite high overall translation scores, localization introduced substantial additional signal. This variation across languages highlights the interaction between translation quality, cultural entity coverage, and the benefits of localization.

\bigskip
\FloatBarrier

\begin{figure*}[!h]
\centering
\includegraphics[width=0.93\textwidth]{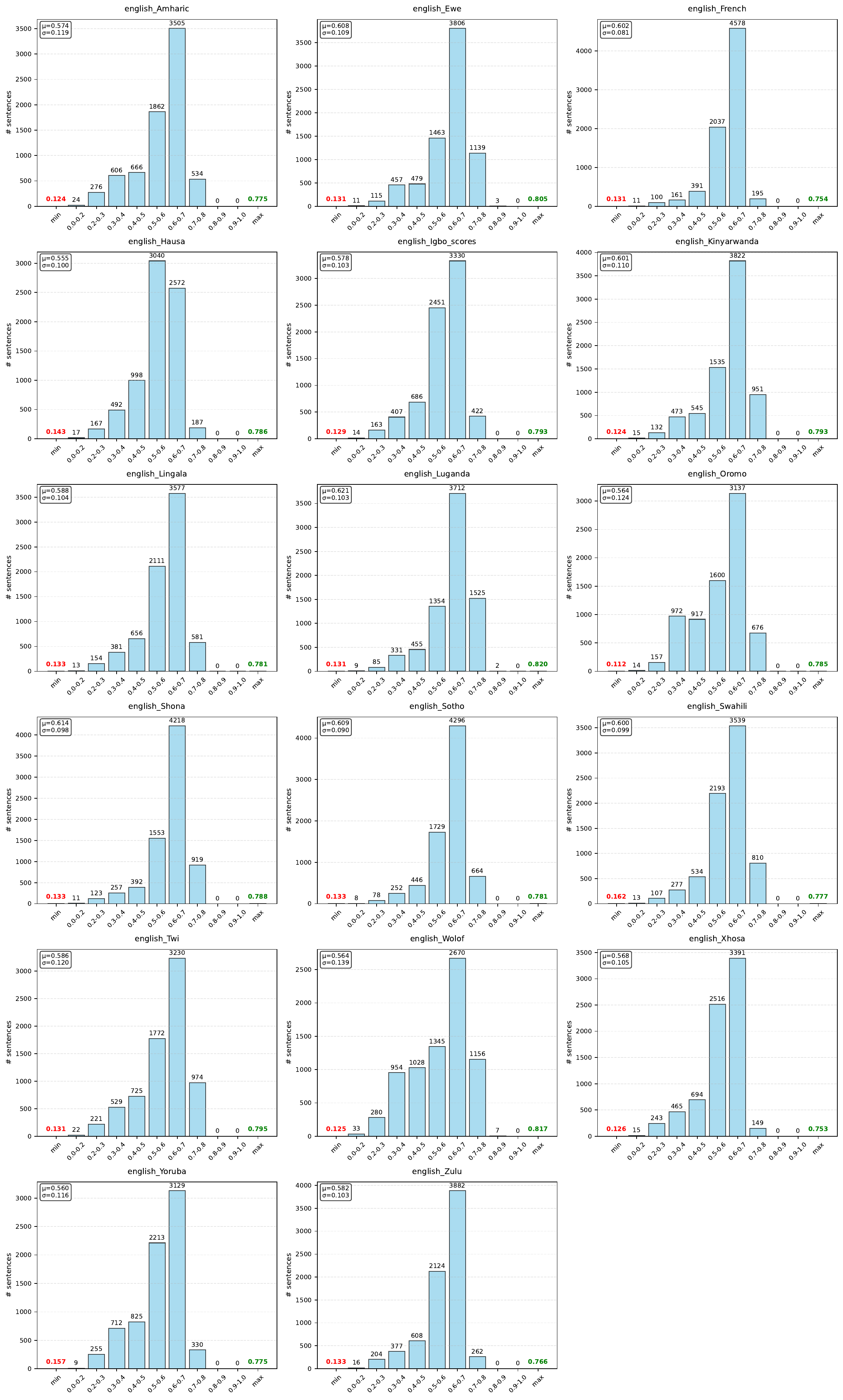}
\caption{Comet score caption here, Score ranges from 0-1, 1 is high translation quality}
\end{figure*}

\section{Error Analysis on RQ2: Native Language Outputs}
Table \ref{error} shows comparison of numeric answers versus non-numeric answers across prompts for different models. Because of this issue in the results making fair comparison challenging, we reported the best result across prompts instead of averaging them for Figure \ref{native_robustness_in_native_language}, similar to prior work \cite{kamruzzaman2025anger}. The results are for 17 languages, 3 prompts, 3 experiments and 250 data each.

\begin{table}[!ht]
\centering

\scalebox{0.}{
\begin{tabular}{cc}
\begin{tabular}{lrr}
\toprule
 & \multicolumn{2}{c}{\llamasit} \\
 & Numeric  & Non Numeric  \\
\midrule
prompt\_1 & 12741 & 9 \\
prompt\_2 & 12369 & 381 \\
prompt\_3 & 11609 & 1141 \\
\bottomrule
\end{tabular}
&
\begin{tabular}{lrr}
\toprule
 & \multicolumn{2}{c}{\gptfmini} \\
 & Numeric  &  Non Numeric \\
\midrule
prompt\_1 & 12734 & 16 \\
prompt\_2 & 12725 & 25 \\
prompt\_3 & 12670 & 80 \\
\bottomrule
\end{tabular}
\\[1em]
\begin{tabular}{lrr}
\toprule
 & \multicolumn{2}{c}{\cohere} \\
 & Numeric & Non Numeric \\
\midrule
prompt_1 & 12708 & 42 \\
prompt_2 & 5469  & 7281 \\
prompt_3 & 12286 & 464 \\
\bottomrule
\end{tabular}
&
\begin{tabular}{lrr}
\toprule
 & \multicolumn{2}{c}{\gemmatwentyit} \\
 & Numeric & Non Numeric \\
\midrule
prompt_1 & 12750 & 0 \\
prompt_2 & 12737 & 13 \\
prompt_3 & 12734 & 16 \\
\bottomrule
\end{tabular}
\end{tabular}
}
\caption{Error analysis on RQ2 (Native language outputs) of selected models. }
\label{error}
\end{table}

\section{Code and Reproducibility}
We used ElutherAI's open source Language Model Evaluation Harness (lm-eval) framework~\cite{eval-harness} to evaluate models. Instead of direct model querying this allows us to have standardized reproducible evaluations across all LLMs. We also set generation parameters (i.e. temperature) to zero for consistency.

Due to the challenges of obtaining numeric answers in low-resource mathematical evaluations, we adopted a strategy of extracting the best answer option and reporting numeric match scores based on the best-performing prompt for each model (Table \ref{native_robustness_in_native_language}). In contrast, since we were able to obtain a sufficient number of numeric answers for English math word problems (MWPs), we averaged the results across three prompts and report these in Table \ref{eng_with_entities}.





\begin{table*}[!ht]
\centering
\begin{tabular}{p{3cm} p{7cm}}
\hline
\textbf{Section} & \textbf{Parameters} \\
\hline
\textbf{Model} & 
\texttt{model\_name\_or\_path: [google/gemma-2-9b-it, meta-llama/Meta-Llama-3-8B-Instruct]} \newline
\# choose one \\
               & \texttt{trust\_remote\_code: true} \\
\hline
\textbf{Method} & 
\texttt{stage: sft} \newline
\texttt{do\_train: true} \newline
\texttt{finetuning\_type: lora} \newline
\texttt{lora\_rank: 8} \newline
\texttt{lora\_target: all} \\
\hline
\textbf{Dataset} & 
\texttt{dataset: [gsm8k-math-localized, gsm8k-math-english, gsm8k-math-models-translated]} \newline
\# choose one or all \newline
\texttt{template: [gemma2, llama3]} \newline
\# choose one \newline
\texttt{cutoff\_len: 2048} \newline
\texttt{max\_samples}: \#1000,5000,10000,none \newline
\texttt{overwrite\_cache: true} \newline
\texttt{preprocessing\_num\_workers: 16} \newline
\texttt{dataloader\_num\_workers: 4} \\
\hline
\textbf{Output} & 
\texttt{output\_dir: \#some directory} \newline
\texttt{logging\_steps: 10} \newline
\texttt{save\_steps: 500} \newline
\texttt{plot\_loss: true} \newline
\texttt{overwrite\_output\_dir: true} \newline
\texttt{save\_only\_model: false} \newline
\texttt{report\_to: none} \\
\hline
\textbf{Train} & 
\texttt{per\_device\_train\_batch\_size: 1} \newline
\texttt{gradient\_accumulation\_steps: 8} \newline
\texttt{learning\_rate: 1.0e-4} \newline
\texttt{num\_train\_epochs: 3.0} \newline
\texttt{lr\_scheduler\_type: cosine} \newline
\texttt{warmup\_ratio: 0.1} \newline
\texttt{bf16: true} \newline
\texttt{ddp\_timeout: 180000000} \newline
\texttt{resume\_from\_checkpoint: null} \\
\hline
\end{tabular}
\caption{Configuration for fine-tuning \texttt{gemma-2-9b-it} and \texttt{Meta-Llama-3-8B-Instruct}.}\label{tab:gemma-config}
\end{table*}

\end{document}